\documentclass[10pt,twocolumn,letterpaper]{article}

\usepackage{cvpr}              

\usepackage{graphicx}
\usepackage{amsmath}
\usepackage{amssymb}
\usepackage{booktabs}

\usepackage[pagebackref,breaklinks,colorlinks]{hyperref}

\usepackage[capitalize]{cleveref}
\crefname{section}{Sec.}{Secs.}
\Crefname{section}{Section}{Sections}
\Crefname{table}{Table}{Tables}
\crefname{table}{Tab.}{Tabs.}

\usepackage[labelformat=simple]{subcaption}

\usepackage{array}
\usepackage{anyfontsize}

\def\cvprPaperID{*****} 
\def\confName{CVPR}
\def\confYear{2022}

\begin{document}

\title{{\fontsize{14}{16}\selectfont SAMPLE-HD: Simultaneous Action and Motion Planning Learning Environment}}

\author{Michal Nazarczuk \qquad Tony Ng \qquad Krystian Mikolajczyk\\
Imperial College London\\
{\tt\small [michal.nazarczuk17, tony.ng14, k.mikolajczyk]@imperial.ac.uk}
}

\maketitle

\begin{abstract}
    Humans exhibit incredibly high levels of multi-modal understanding - combining visual cues with read, or heard knowledge comes easy to us and allows for very accurate interaction with the surrounding environment. 
    Various simulation environments focus on providing data for tasks related to scene understanding, question answering, space exploration, visual navigation. 
    In this work, we are providing a solution to encompass both, visual and behavioural aspects of simulation in a new environment for learning interactive reasoning in manipulation setup.  SAMPLE-HD environment allows to generate various scenes composed of small household objects, to procedurally generate language instructions for manipulation, and to generate ground truth paths serving as training data. 
\end{abstract}

\section{Introduction}


With the increasing interest in developing  systems \cite{Huang2019TransferableNavigation, Kim2021ViLT:Supervision, Shridhar2020ALFREDTasks} capable of human like perception and reasoning capabilities, the amount of data required to develop data-driven approaches is constantly increasing and various syntheticaly generated datasets were introduced \cite{Anderson2018Vision-and-LanguageEnvironments, Chang2017Matterport3D:Environments, Savva2019Habitat:Research, Wu2018BuildingEnvironment}. Similarly, robotics community frequently uses simulated environments \cite{Fan2018SURREAL:Benchmark, Tassa2018DeepMindSuite} as means of training and testing algorithms due to the high cost of gathering data using the real hardware. However, synthetic environments that offer accurate physics simulation usually suffer from low visual quality \cite{Miller2004GraspIt, Todorov2012MuJoCo:Control}. In this work, we are trying to bridge the gap between vision and robotics by proposing a new environment providing precise physics calculation along high quality visual output in order to facilitate training of  reasoning algorithms  in object manipulation setup.

Robotic manipulators are one of the main topics of interest in the robotics community. However, obtaining data to train neural networks with visual input involves either a time consuming acquisition process (restricted by real time operations) or training with the use of simulated environment. The latter often suffers from unrealistic visual quality making it less interesting from computer vision perspective.

SAMPLE-HD is a simulation environment build in Unity \cite{Unity-TechnologiesUnity3D2020} that provides high quality visual data characterised by realistic physics behaviour. Thus, it can be used for tasks of scene understanding and controlling the robotic manipulator. We focus the simulation on a kitchen tabletop scenario, in which the robotic system is asked to perform object manipulation instructed by a given natural language query. 

 \begin{figure}[t]
     \centering
     \includegraphics[width=0.95\linewidth]{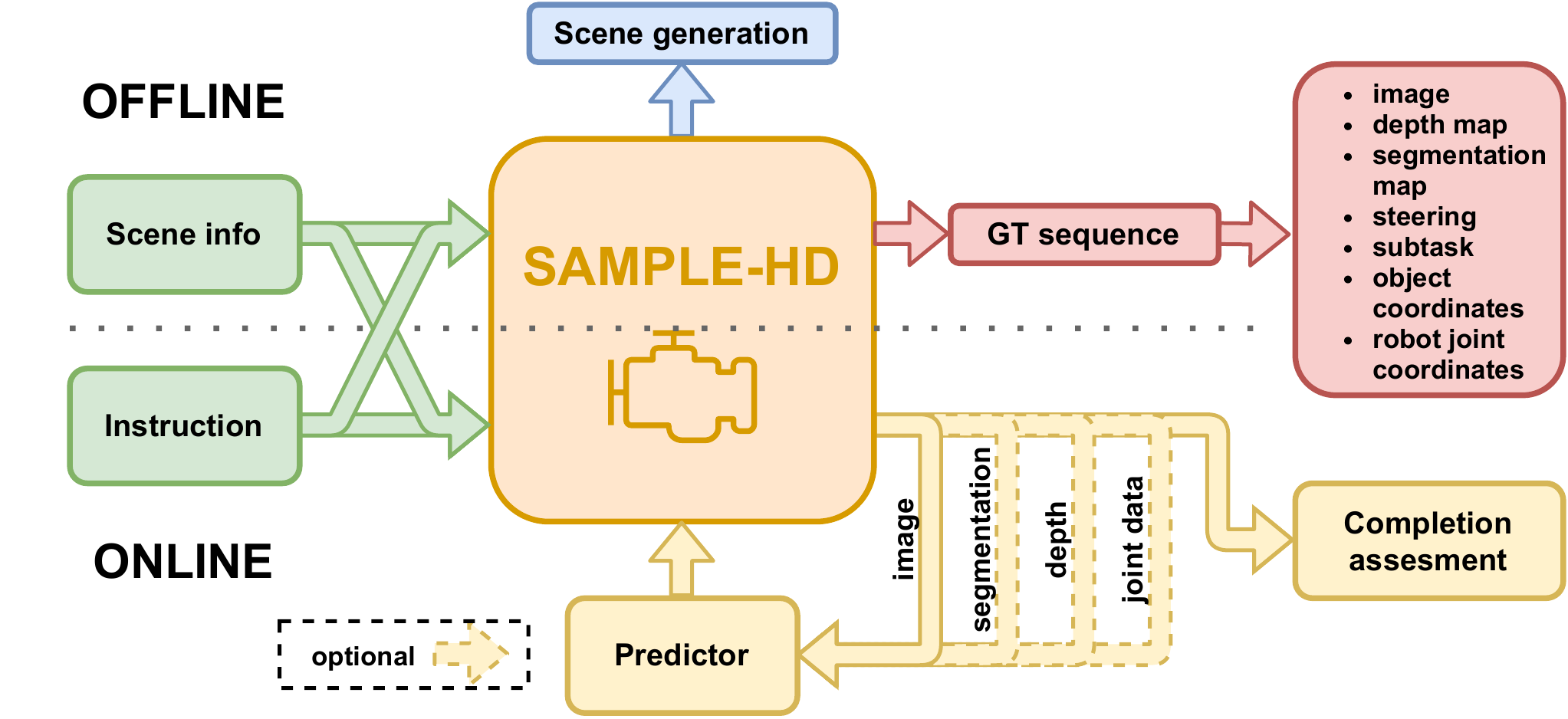}
     \caption{An overview of SAMPLE environment. Dotted line indicates offline and online capabilities. Different outward arrows show different steps of data generation or possible outputs. 
     }\label{fig:overview}
     \vspace{-1em}
 \end{figure}
 
SAMPLE-HD provides a set of utilities to train visual systems to predict a steering signal for a robotic arm. An overview of the environment is presented in Figure \ref{fig:overview}. Firstly, it can generate various scenes, in which kitchen appliances (based on SHOP-VRB \cite{Nazarczuk2020SHOP-VRB:Perception}) are randomly placed on the table with a randomised appearance. Additionally, we provide a way to procedurally generate a set of instructions asking the system to perform various interactions with the scene. The offline generated training data  includes also a sequence of ground truth images and steering values required  to fulfil the task. For the inference, or online training, we utilise ML-Agents \cite{Juliani2020Unity:Agents} toolkit providing an API to facilitate communication between environment and the network.

Due to its modular nature, SAMPLE-HD allows for easy modifications. All the models used can be swapped for any imported 3D model which can conveniently be customised with respect to randomisation options. With the use of constrained-based templates, modifications to the instruction generator are also viable. Finally, ground truth manipulator paths are also based on simple modules allowing to stack them differently in order to obtain desired result.  

 \begin{figure*}[t]
     \centering
     \includegraphics[width=0.98\linewidth]{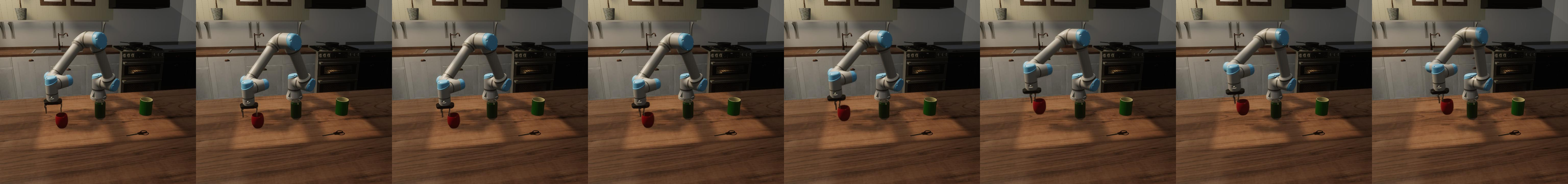}
     \caption{Part of a GT sequence generated in SAMPLE-HD. Images are cropped, and sampled every $14th$ frame for better presentation.}\label{fig:sequence}
 \end{figure*}

\section{Environment}



Scenes in SAMPLE-HD are placed in virtual kitchen environment. 
All the objects and the manipulator are placed on the table. Robotic arm chosen for the simulation is Universal Robots UR5e
equipped with a parallel gripper. 

\noindent
\textbf{Steering} of the robotic arm can be realised in different modes: in joint coordinates or cartesian space
realised with inverse kinematics model.
The control of the robotic arm is implemented asynchronously to physics calculation 
allowing to simulate vision-based decision making while preserving continuous movement. 
Self-collisions, collisions between robot and tabletop, and excessive forces are triggering safety mechanism stopping the arm movement.

\noindent
\textbf{Scene generation} in SAMPLE-HD is fully randomised. Given the set of assets, each scene is generated via subsequent placement of objects in the scene. Every object is chosen randomly and its properties are also randomised. Via a menu attached to the object prototype, the user can choose which properties, among colour, material, size, of the asset can be changed. When placing the object, the possibility of intersections and occlusions with other objects are checked
and incorrect positions are rejected. Finally, the scene is generated and rendered with the view from randomly positioned camera, segmentation map, depth map, ground truth objects data (in the manner of CLEVR annotations), and manipulator position.



\noindent
\textbf{Instructions} asking the robotic arm to perform various actions related to manipulations are generated on the base of templates similar to CLEVR. We have noticed that when using depth first search on all possible adjectives combinations (like in CLEVR) leads to high execution times due to exponential increase with the number of adjectives. Therefore, we propose a rejection based generation of the instructions. Firstly, a random template is chosen assuring uniform distribution of either templates or resulting task types. For a chosen template and scene we generate a set of object level descriptors, depending on the required type - unique or non-unique. Descriptors are generated to provide a shortest unique (non-) description of the scene objects. In such way a more natural descriptions are created - it is more convenient to say \textit{blue bowl} instead of \textit{light, medium-sized, blue, hemispherical bowl}. Thereafter, we perform constraint based rejection test in order to filter out descriptions related to objects not fulfilling template criteria. Finally, filtered combinations are used to randomly choose set of descriptions and generate instructions. Additionally, we provide CLEVR-compatible programs for each instruction that can be used for symbolic approaches like \cite{Johnson2017InferringReasoning, Yi2018Neural-SymbolicUnderstanding}. Finally, a descriptor of task along with pointers to target scene objects are included.

\noindent
\textbf{Motion paths generation} is provided to obtain motion paths for generated instructions along with corresponding values of steering that may serve as the supervision for various data-driven approaches. Given the scene and the corresponding instruction, the task is split into major movements, namely: approaching the object's neighbourhood, assuming grasping position, performing grasping, movement with object to destination, releasing the object. 

\noindent
\textbf{Dataset} utilising SAMPLE-HD is provided along the environment. We have chosen to use objects from SHOP-VRB. 
We provide $2000$ scenes containing $3$ to $5$ objects. 
For each scene we attempt to generate instructions from provided templates resulting in $8360$ instructions. 
Finally, a ground truth motion path was recorded at every rendered frame. The output consists of images rendered by a camera, segmentation and depth maps, robot position in joint coordinates, motion subtask and steering vector, and positions and orientations of every object in the scene (in such way the whole scene can be reconstructed if necessary). Additionaly, SAMPLE-HD is equipped with a tool for assessing whether the task was fulfilled successfully. Example part of generated sequence is shown in Figure \ref{fig:sequence}. 
 As a part of dataset, we generated ground truth paths for all $8360$ instructions that resulted in providing nearly $4.5mln$  data points for training.

\begin{figure}[h]
    \centering
    \begin{minipage}{0.61\linewidth}
        \centering
        \includegraphics[width=0.99\textwidth]{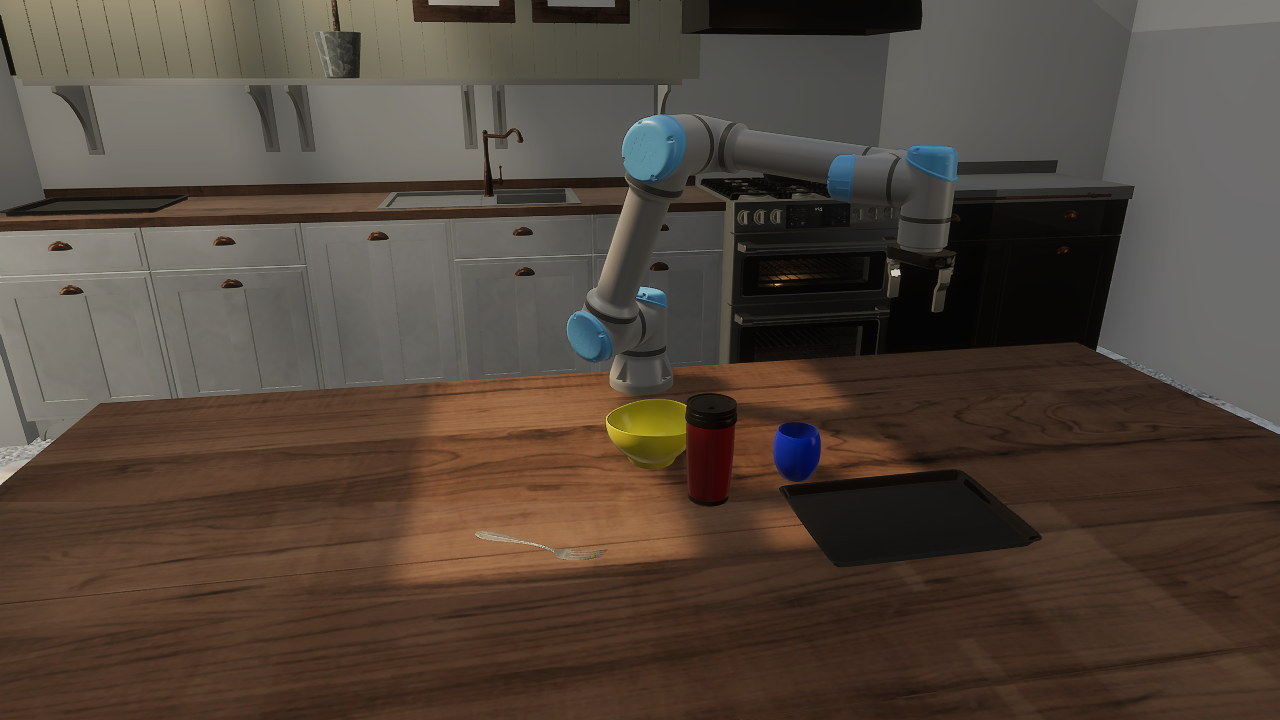}
        \vspace{1mm}
        \begin{minipage}{0.975\textwidth}
            {\scriptsize Stack blue glass on top of metal object that is right to the thermos.}
        \end{minipage}
    \end{minipage}
    \begin{minipage}{0.38\linewidth}
        \includegraphics[width=0.99\textwidth]{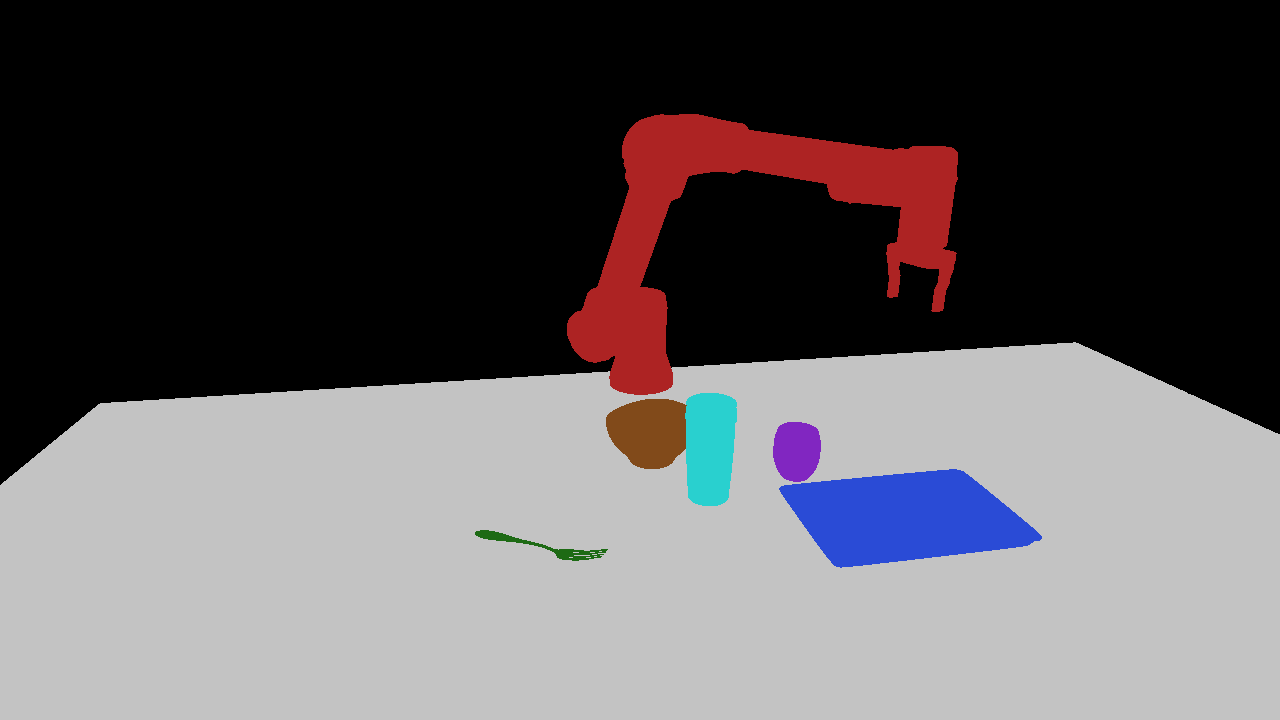}
        \includegraphics[width=0.99\textwidth]{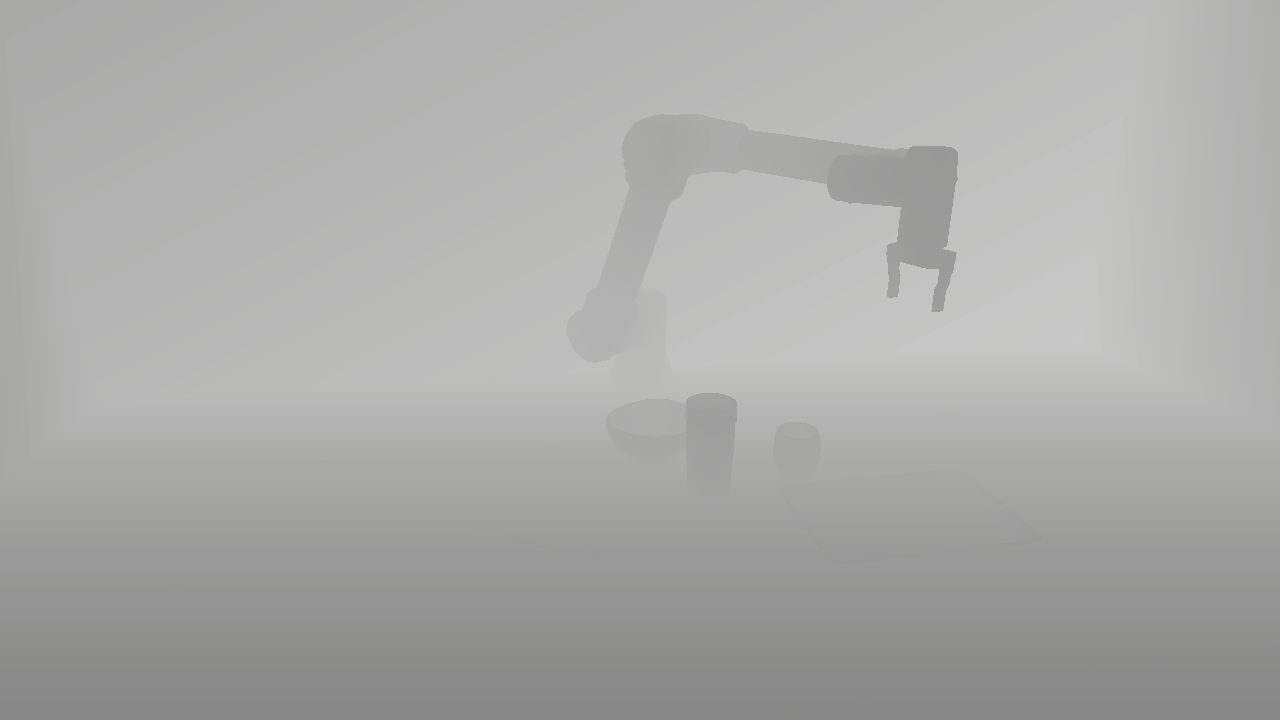}
    \end{minipage}
    \caption{Example scene (left) from proposed dataset along with generated segmentation map (top right), depth map (bottom right), and sample instruction (bottom right).} \label{fig:scene_example}
\end{figure} 
{\small
\bibliographystyle{ieee_fullname}
\bibliography{egbib}
}

\end{document}